\pdfoutput=1
\documentclass[11pt]{article}

\usepackage[final]{acl}

\usepackage{times}
\usepackage{latexsym}
\usepackage{amsmath}
\usepackage{subcaption}
\usepackage{booktabs}
\usepackage{colortbl}
\usepackage[T1]{fontenc}
\usepackage[utf8]{inputenc}

\usepackage{microtype}

\usepackage{inconsolata}

\usepackage{makecell}
\usepackage[table,dvipsnames]{xcolor}
\usepackage[table]{xcolor}
\usepackage{graphicx}
\usepackage{algorithm}
\usepackage{algpseudocode}
\usepackage{tabularx}
\usepackage{float}
\usepackage{placeins}
\usepackage{caption}
\usepackage{enumitem}
\usepackage{hyperref}  % Required for \autoref

\usepackage{amsmath}
\usepackage{graphicx}   % 이미지 삽입
\usepackage{lipsum} % 아무말 텍스트 삽입 보기위한 패키지
\usepackage{cuted} % 2column 1column 혼용
\usepackage{stfloats}
\usepackage{multicol}
\usepackage{color}
\usepackage[normalem]{ulem}
\usepackage{amsfonts} % mathbb 수학 수식 폰트
\usepackage{colortbl} % 표 배경색
\usepackage{multirow} % 표 다중행 병합
\usepackage{tcolorbox}
\usepackage{booktabs}
\usepackage[table]{xcolor}
\usepackage{arydshln}  
\title{\textsc{Steam}: A Semantic-Level Knowledge Editing Framework\\
for Large Language Models}

\author{Geunyeong Jeong,\; Juoh Sun,\; Seonghee Lee,\; Harksoo Kim\thanks{Corresponding author.}\\
Konkuk University\\
\texttt{\{jyjg7218, qssz1326, nlpshlee, nlpdrkim\}@konkuk.ac.kr}}

\begin{document}
\maketitle
\begin{abstract}
Large Language Models store extensive factual knowledge acquired during large-scale pre-training. However, this knowledge is inherently static, reflecting only the state of the world at the time of training. Knowledge editing has emerged as a promising solution for updating outdated or incorrect facts without full retraining. However, most existing locate-and-edit methods primarily focus on token-level likelihood optimization without addressing semantic coherence. Our analysis reveals that such edited knowledge is often encoded as isolated residual streams in the model's latent space, distinct from pre-existing knowledge and bypassing natural reasoning process. To address this, we propose \textsc{Steam}, a semantic-level knowledge editing framework that enhances integration of updated knowledge into the model's knowledge structure. \textsc{Steam} first identifies target representations as semantic anchors for the updated factual association, then guides the internal representation of the edited fact towards these anchors through an alignment loss during optimization. Experimental results demonstrate that \textsc{Steam} improves model’s ability to reason with edited knowledge and enhances semantic coherence, underscoring the importance of latent-space alignment for reliable and coherent knowledge editing. The code is available at \url{https://github.com/GY-Jeong/STEAM}.
\end{abstract}

\section{Introduction}
Large Language Models (LLMs) have achieved impressive performance in knowledge-intensive NLP tasks~\cite{ Knowledge_intensive_2,mmlu_pro} by leveraging extensive factual knowledge acquired through large-scale pre-training~\cite{toward_tracing,KnowledgeMechnism,chang2024large}. However, this knowledge is inherently static, reflecting only the state of the world at the time of training. This limitation necessitates efficient methods for updating LLMs when new information arises or existing facts become outdated.

Knowledge editing\footnotemark{}~\cite{edit_factual_LLM, layer_wise_info, editing_survey} has emerged as a promising approach, enabling selective updates to specific factual information without the need for costly full retraining. This approach typically identifies a fact as a triple structure $(s,r,o)$, representing (subject, relation, object), and aims to replace the original object $o$ with a new entity $o^*$. For example, the outdated fact (\texttt{UK}, \texttt{Prime Minister}, \texttt{Rishi Sunak}) can be updated to the current fact (\texttt{UK}, \texttt{Prime Minister}, \texttt{Keir Starmer}).

\footnotetext{%
  In this paper, we focus on parameter-modifying knowledge editing methods,
  specifically locate-and-edit approaches.%
}
For knowledge editing to be reliable, it is crucial that updated facts are consistently integrated into the model’s knowledge structure. However, most existing methods primarily focus on token-level likelihood optimization (i.e., maximizing the generation probability of $o^*$) without addressing semantic coherence. Consequently, these methods face challenges in tasks that require knowledge integration, such as maintaining consistency across interconnected facts or supporting complex reasoning tasks~\cite{CounterFactPlus, Ripple_effect, ClozePrompt}.

To investigate this limitation, we examine how edited knowledge is represented within the latent semantic space of the edited model. Our analysis reveals that \textbf{edited knowledge is often encoded as isolated residual streams}\,--\,sequences of hidden states propagated through the transformer layers\,--\,which are distinct from model's pre-existing knowledge (\autoref{sec:3_2}). Building on this finding, additional interpretability-based analysis reveals that \textbf{these residual streams reflect a direct activation for the generation of the target token $o^*$}, bypassing the model’s natural reasoning process (\autoref{sec:3_3}). This behavior exposes a fundamental shortcoming of current editing strategies and underscores the need for approaches that promote semantic-level integration for coherent and reliable knowledge editing.

To address this challenge, we propose \textsc{Steam} ({\textbf{S}eman\textbf{t}ic-Level Knowledge \textbf{E}diting Fr\textbf{am}ework}), a novel approach that enhances knowledge integration and is easily compatible with existing locate-and-edit approaches. \textsc{Steam} augments the conventional locate-and-edit approach through two main components: (1) \textbf{Latent Positioning}, which identifies semantic anchors for edited knowledge, and (2) \textbf{Latent-Level Alignment}, which guides the edited knowledge representation towards these anchors in the latent space. Our experimental results demonstrate that \textsc{Steam} significantly improves the model's reasoning ability with edited knowledge and enhances overall semantic consistency, supporting the advantages of semantic-level integration for reliable knowledge editing.
\newline \newline \noindent We summarize our contributions as follows:
\begin{itemize}[leftmargin=8pt]
\setlength\itemsep{0em}
    \item We systematically analyze conventional locate-and-edit methods and reveal that they encode edited knowledge separately from the model's existing knowledge (\autoref{sec:3}).
    \item We propose \textsc{Steam}, a semantic-level knowledge editing framework designed to enhance the semantic coherence of updated knowledge with the model's internal knowledge representations (\autoref{sec:4}).
    \item We empirically demonstrate that \textsc{Steam} significantly improves reasoning with edited facts and enhances semantic coherence, across diverse baselines and editing settings (\autoref{sec:5}).
\end{itemize}

\section{Locate-and-Edit Approach}\label{sec:2}
Locate-and-edit approach consist of two stages: (1) \textbf{Locating}, which identifies the parameters where a fact $(s,r,o)$ is stored, and (2) \textbf{Editing}, which modifies those parameters to encode the updated fact $(s,r,o^*)$.

\subsection{Locating}\label{sec:2_1}
To identify the parameters responsible for encoding factual associations (i.e., $(s,r)\rightarrow o$) within LLMs, prior work has employed knowledge analysis techniques such as causal tracing~\cite{ROME}. These methods systematically perturb intermediate model representations to pinpoint the components that contribute to factual recall. Empirical findings suggest that factual associations are primarily encoded in the MLP modules of the early-to-middle transformer layers~\cite{meng2023massediting}.

\subsection{Editing}\label{sec:2_2}
Previous studies have identified MLP modules within LLMs function as key-value memory structures~\cite{key-value_memory}. Specifically, the first layer $W_{fc}$ functions as a key encoder, while the second layer $W_{proj}$ retrieves the corresponding value. In this context, the final subword of the subject $s$ serves as the key $k$, and the relation–object pair $(r,o)$ is represented as the value $v$.

As the first step of the editing phase, the updated fact $(s, r, o^*)$ is encoded into a new key-value pair $(k^*, v^*)$. The key $k^*$ is computed by averaging hidden states at the final subword position of $s$ across multiple contexts augmented with prefix prompts $x_j$. The value $v^*$ is obtained through an iterative optimization process that minimizes the following composite objective:
\begin{flalign}
    \min_{\delta}\mathcal{L}(\delta)=\mathcal{L}_{\mathrm{NLL}}(\delta)+\mathcal{L}_{\mathrm{KL}}(\delta),
\label{eq:obj}
\end{flalign}
where $\delta$ denotes a candidate vector for $v^*$. The first term, $\mathcal{L}_{\mathrm{NLL}}(\delta)$, maximizes the likelihood of generating $o^*$ given a factual prompt $p$ constructed from $(s,r)$:
\begin{flalign}
    \mathcal{L}_{\mathrm{NLL}}(\delta)=-\frac{1}{N}\sum_{j=1}^{N}\log \mathbb{P}(o^* \mid x_j + p\ ;\,\delta),
\label{eq:nll}
\end{flalign}
where $\mathbb{P}(\cdot)$ denotes the token generation probability of the model. The second term, $\mathcal{L}_{\mathrm{KL}}(\delta)$, minimizes semantic drift by reducing the KL divergence between the updated and original predictions for a generic prompt $p'$ (e.g., “{subject} is a”):
\begin{flalign}
\mathcal{L}_{\mathrm{KL}}(\delta)=D_{\mathrm{KL}}(\mathbb{P}(x \mid p';\,\delta)\,\|\,\mathbb{P}(x \mid p')).
\label{eq:kl}
\end{flalign}

To incorporate the updated key–value pair ($k^*$, $v^*$) into the model, the original projection matrix $W_{proj}$ (denoted as $W$ for simplicity) is modified to produce an updated weight matrix $\hat{W}$. This update is defined as:
\begin{equation}
    \begin{aligned}
        \hat{W}=W+\Lambda(C^{-1}k^*)^T,\\
    \text{where}\quad\Lambda=\frac{v^*-Wk^*}{(C^{-1}k^*)^T k^*}.
    \end{aligned}
    \label{eq:edit_equation}
\end{equation}
Here, $C = KK^T$ captures the second-order statistics of existing keys $K$, which encode preserved knowledge within the model. The edited model $\mathcal{F'}$ is obtained by replacing the original matrix $W$ with the updated matrix $\hat{W}$, thereby incorporating the new factual association into the model.

While this minimal perturbation enables effective knowledge updates, it operates by optimizing token-level generation probabilities (Eq.~\ref{eq:nll}). This raises the question of how the model understands the edited knowledge: \textbf{Is it hard-coded for output prediction or integrated into the model’s knowledge structure?} In the following analysis, we address this question by analyzing how edited facts are represented and processed within the model’s latent space.

\section{Latent Space Analysis of Edited Knowledge}\label{sec:3}
Although an edited language model can generate the updated fact, this alone does not confirm its integration into the model’s knowledge structure. To investigate this, we conduct two complementary analyses focusing on the residual stream, the sequence of hidden states propagated through the transformer layers:
\begin{itemize}[leftmargin=8pt, topsep=6pt]
\setlength\itemsep{0em}
    \item We visualize how edited knowledge propagates through the model’s latent space (\autoref{sec:3_2}).
    \item We interpret the semantic implications of these residual streams using the LogitLens technique (\autoref{sec:3_3}).
\end{itemize}

For this analysis, we sample 1,000 instances from the \textsc{CounterFact}~\cite{ROME} dataset, specifically focusing on cases where the factual association $(s,r) \rightarrow o$ is preserved in the model, and both the original target $o$ and the edited target $o^*$ are single-token entities.

\subsection{Extracting Layer-wise Representations} \label{sec:3_1}
\begin{figure*}[t]
  \includegraphics[width=\linewidth]{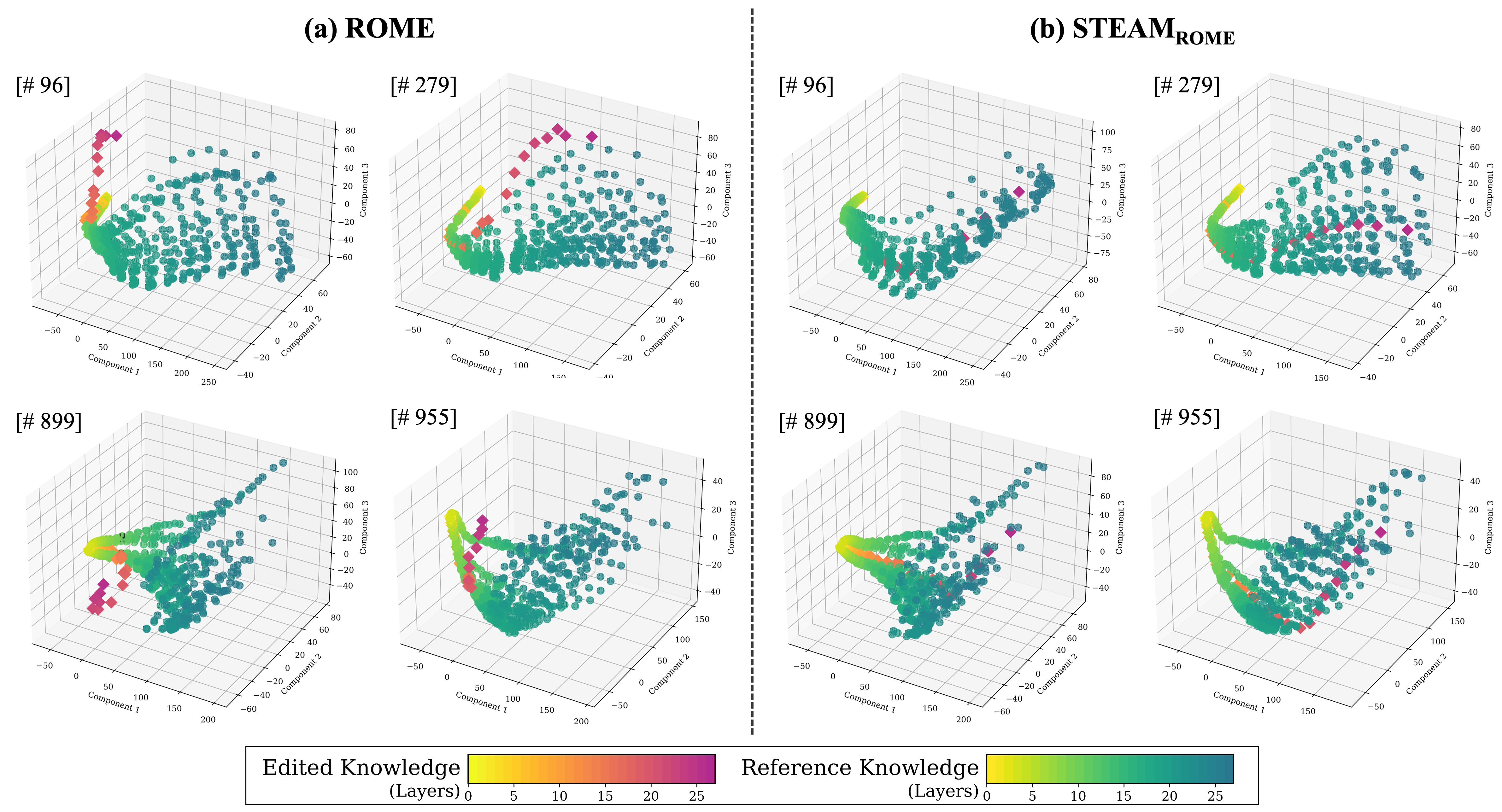}
  \caption{3D visualization of residual stream across layers. Each subplot shows PCA-projected hidden states for a sample from the \textsc{CounterFact} dataset (index in brackets). Red diamonds represent the residual stream of the edited knowledge, while green circles denote that of the reference knowledge. (a) shows the result after applying ROME, and (b) shows the result with \textsc{Steam}$_\text{ROME}$.}
  \label{fig:overview_rome_vs_stake}
\end{figure*}
We first establish a baseline for how the base model encodes the target object $o^*$ given diverse factual contexts. For a given edit $\varepsilon$ : $(s, r, o \rightarrow o^*)$, we collect a set of reference triples $T=\{(s_i, r_i, o^*)\}_{i=1}^N$ sourced from Wikidata\footnote{\url{https://www.wikidata.org}}, a structured knowledge base. For instance, given an edit $\varepsilon$ : (\texttt{Eiffel Tower}, \texttt{located in}, \texttt{Paris} → \texttt{London}), the set $T$ may include related triples such as (\texttt{River Thames}, \texttt{flows through}, \texttt{London}) and (\texttt{Big Ben}, \texttt{located in}, \texttt{London}). Since the model may not have internalized all facts in $T$, we filter out triples for which the model fails to accurately recall $o^*$. Following prior work~\cite{ClozePrompt}, we construct a cloze-style textual prompt $p_i$ for each pair $(s_i, r_i)$ and feed it into the model $\mathcal{F}$. For example, given (\texttt{River Thames}, \texttt{flows through}), we use the prompt “\emph{River Thames flows through \underline{\hspace{0.5cm}}}”. Through this filtering and sampling process, we construct $T' \subseteq T$ , a subset of reference knowledge that the model has internalized.

For each prompt $p_i \in T'$, we extract the layer-wise hidden representations  $\{h_i^\ell\}_{\ell=1}^L$ from all $L$ layers of the base model $\mathcal{F}$, at the final token position where $o^*$ is predicted. This captures how the model represents pre-existing knowledge. Similarly, for the edited model $\mathcal{F'}$, we input the edited prompt $p_\varepsilon$ (e.g., “\emph{Eiffel Tower is located in \underline{\hspace{0.5cm}}}”) and extract its hidden states $\{h_\varepsilon^\ell\}_{\ell=1}^L$ at the corresponding prediction position, reflecting how the model encodes the edited fact.

\subsection{Layer-wise Visualization of Edited Knowledge Flow} \label{sec:3_2}
\begin{tcolorbox}[
    colback=gray!10,  % Background color (light gray)
    colframe=gray!10, % Frame color (same as background for no border)
    boxrule=0pt,      % No border
    sharp corners,    % No rounded corners
    width=\columnwidth % Match text width
]
\textbf{Takeaway 1:} Edited knowledge is represented as a isolated residual stream in the latent space, indicating a fundamental issue with semantic integration.
\end{tcolorbox}
\noindent To better understand how edited knowledge is represented and propagated within the model, we conduct a qualitative analysis based on residual stream visualization. Specifically, we compare the layer-wise hidden representations of the edited knowledge $\{h_\varepsilon^\ell\}_{\ell=1}^L$ and the reference knowledge $\{h_i^\ell\}_{\ell=1}^L$ to examine differences in how their semantic representations evolve across layers. We begin by applying Principal Component Analysis (PCA)~\cite{PCA} to reduce the dimensionality of the hidden states. In Figure~\ref{fig:overview_rome_vs_stake}(a), we plot the hidden states corresponding to edited fact as red diamonds, and those corresponding to the reference knowledge as green circles. Additional visualizations for other samples are provided in Appendix~\ref{sec:appendix_visualization}.

As shown in Figure~\ref{fig:overview_rome_vs_stake}(a), there is a clear distinction between edited and reference knowledge in the model's latent space. The representations of reference knowledge evolve progressively across layers, becoming more dispersed in the deeper layers, which reflects their gradual integration into broader semantic contexts~\cite{ContextInformation_layerwise}. In contrast, the representations of edited knowledge follow a separate and isolated path. This divergence suggests that the model encodes edited knowledge independently, rather than incorporating it into the its broader knowledge structure.

Notably, this separation is apparent from the mid-layers, which are crucial for constructing knowledge associations~\cite{Dissecting, ContextInformation_layerwise}. This observation raises concerns that the edited factual association is formed in a manner inconsistent with the preserved knowledge. Since LLMs organize knowledge in an entity-centric manner, similar to structured knowledge bases~\cite{LLM_knowledge_base_1, LLM_knowledge_base_2}, this discrepancy potentially leads to difficulties in leveraging related facts, thereby hindering complex reasoning.

To complement the qualitative findings, we additionally perform a quantitative analysis using cosine similarity between hidden states of edited and reference knowledge. This supplementary analysis supports our observations and is presented in Appendix~\ref{appendix:cosine_sim_analysis}.

\subsection{Interpreting Reasoning Process of Edited Knowledge} \label{sec:3_3}
\begin{figure}[t]
\includegraphics[width=\linewidth]{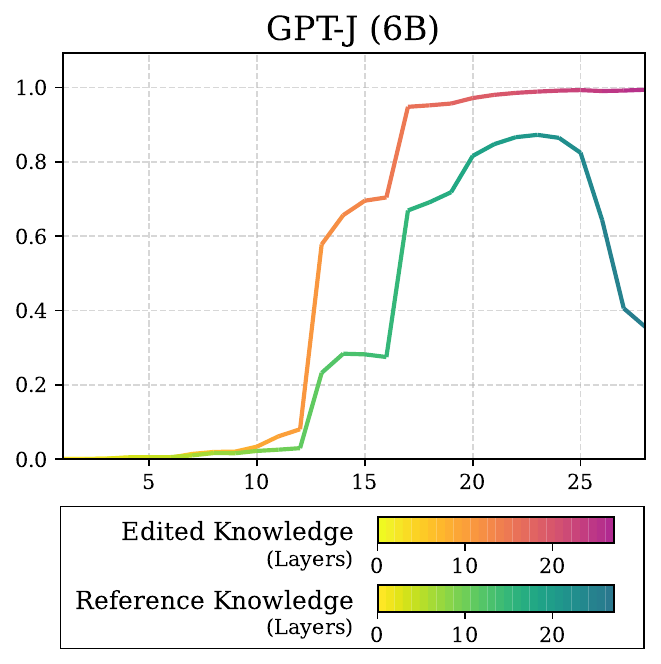}
  \caption{Layer-wise probability of generating the updated object $o^*$ in edited model $\mathcal{F'}$ as computed by LogitLens. The $x$-axis indicates the transformer layer; $y$-axis shows the predicted probability for $o^*$. Results are averaged over all samples in \textsc{CounterFact}.}
  \label{fig:token_prob_wise_layer}
\end{figure}
\begin{tcolorbox}[
    colback=gray!10,  % Background color (light gray)
    colframe=gray!10, % Frame color (same as background for no border)
    boxrule=0pt,      % No border
    sharp corners,    % No rounded corners
    width=\columnwidth % Match text width
]
\textbf{Takeaway 2:} Edited knowledge triggers shortcut-like activations that prioritize the target token generation, bypassing the model’s natural reasoning process.
\end{tcolorbox}
\noindent To further investigate the distinction identified in the previous analysis (\autoref{sec:3_2}), we conduct an interpretability-based analysis. Specifically, we employ LogitLens~\cite{LogitLens}, a technique that approximates the token probability distribution at each transformer layer based on its hidden state. This approach enables us to trace how the model incrementally infers the edited knowledge (i.e., the object $o^*$) as information flows through the layers. Formally, given a hidden state $h \in \mathbb{R}^d$ at a particular layer, LogitLens applies the final projection matrix $W_U \in \mathbb{R}^{|V| \times d}$ followed by a softmax over the vocabulary space $V$:
\begin{equation}
\text{LogitLens}(h) = \text{softmax}(W_Uh) \in \mathbb{R}^{|V|}.
\end{equation}
We compute the layer-wise probability of generating $o^*$ using two sets of hidden states : (1) $ \{h_\varepsilon^\ell\}_{\ell=1}^L$, which represent the reasoning process over the edited fact, and (2) $ \{h_i^\ell\}_{\ell=1}^L$, which correspond to the reasoning process over the reference knowledge.
The results in Figure~\ref{fig:token_prob_wise_layer} show clear differences in how the model processes edited knowledge (red line) versus reference knowledge (green line). For reference knowledge, the probability of predicting the target token increases gradually across layers, reflecting a progressive, context-driven reasoning process. In contrast, for edited knowledge, the probability rises sharply from the mid-layers and quickly saturates. 

This contrast suggests that the model does not infer edited facts through its natural knowledge reasoning process. Instead, it relies on direct activation, particularly from the mid-layers, which acts as a shortcut to maximize the generation probability of the target token $o^*$. This behavior is a direct consequence of current editing strategies prioritizing token-level optimization, highlighting the need for approaches that achieve deeper semantic integration.

\section{Method}\label{sec:4}
\begin{figure*}[t]
\includegraphics[width=\linewidth]{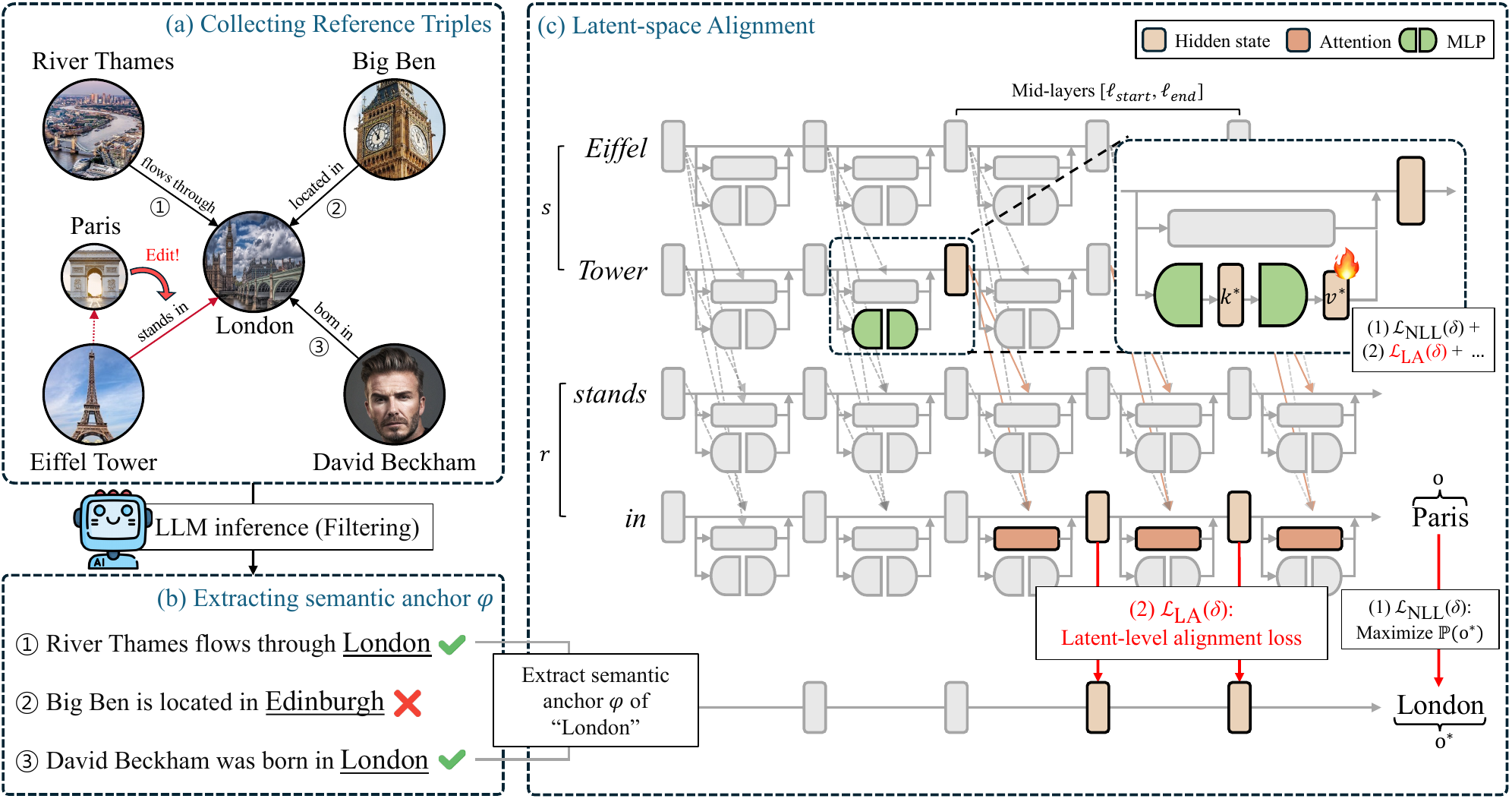}
  \caption{Overview of the \textsc{Steam} framework. (a) Relevant reference knowledge about the new target object $o^*$ (b) The model is used to verify and filter these facts; valid references are then used to construct the semantic anchor $\varphi$ that approximates the latent representation of $o^*$. (c) During editing, \textsc{Steam} introduces a latent-level alignment loss $\mathcal{L}_\text{LA}$, which guides the edited value vector $v^*$ to align with the semantic anchor across mid-layers, encouraging coherent integration of the new knowledge into the model’s latent space.}
  \label{fig:STAKE}
\end{figure*}
To address the semantic isolation of edited knowledge observed in Section~\ref{sec:3}, we introduce \textsc{Steam}, a framework designed to enhance the semantic coherence of updated knowledge with the model's internal knowledge representations. To achieve this, \textsc{Steam} augments the standard locate-and-edit process, specifically enhancing the optimization of the value vector $v^*$ through two main components: (1) \textbf{Latent Positioning}, which identifies appropriate semantic anchors for edited knowledge in the model's latent space, and (2) \textbf{Latent-Level Alignment}, which incorporates an additional objective to steer the edited knowledge representation towards these anchors during the optimization process.

% : Identifying Semantic Anchors for Knowledge Association
\subsection{Latent Positioning}\label{sec:4_1}
The Latent Positioning step aims to identify target representations for the updated factual association $(s,r\rightarrow o^*)$, which we refer to as \textit{semantic anchors}. These anchors approximate how the model would naturally encode the new object $o^*$ in the context of the subject–relation pair $(s, r)$, assuming the fact had been acquired through the model’s standard knowledge acquisition process (e.g., pre-training).

To construct these anchors for a given edit $\varepsilon:(s,r,o\rightarrow o^*)$, we start from the set of reference knowledge $T'$ obtained through the retrieval and filtering process described in Section~\ref{sec:3_1}. From this filtered set, we sample a balanced subset, denoted as $T''$, to mitigate relation-type biases and ensure the anchors reflect diverse relational contexts (details in Appendix~\ref{sec:appendix_reference_triple}). For each prompt $p_i\in T''$, we extract the layer-wise hidden states $ \{h_i^\ell\}_{\ell=1}^L$ from the base model at the final token position corresponding to $o^*$. These hidden states represent how the model internally encodes the target object $o^*$ when processing various subject–relation pairs $(s_i, r_i)$. To obtain a generalized representation that abstracts away from specific contexts and captures the consistent semantics of $o^*$, we compute the mean of these hidden states at each layer:
\begin{equation}
\varphi^\ell=\frac{1}{|T''|} \sum_{i=1}^{|T''|} h^{\ell}_i,\quad\ell = 1,\dots,L.
\end{equation}
The resulting set of vectors $ \{\varphi^\ell\}_{\ell=1}^L$ serves as our semantic anchors, representing context-independent features of $o^*$ derived from diverse, faithfully recalled factual contexts. In our framework, we focus on semantic anchors $\varphi^\ell$ from the mid-layers. This choice is motivated by prior findings that these layers play a central role in encoding relational semantics and attribute-level information~\cite{Relation_Mapping, Dissecting}, making them suitable locus for integrating new factual knowledge in a manner consistent with the model's knowledge processing mechanism.

\subsection{Latent-Level Alignment }\label{sec:4_2}
The Latent-Level Alignment step guides the internal representation of the edited fact $(s,r,o^*)$ towards its corresponding semantic anchors $\varphi$ within the model's latent space. In the conventional locate-and-edit framework (as described in \autoref{sec:2}), a new factual association $(s,r\rightarrow o^*)$ is encoded into a key-value pair $(k^*, v^*)$ within an MLP layer, where the value vector $v^*$ encodes the information of the updated object $o^*$. The optimization of $v^*$ is typically guided by two objectives: the negative log-likelihood loss $\mathcal{L}_{\mathrm{NLL}}$ and the KL divergence loss $\mathcal{L}_{\mathrm{KL}}$. While these objectives  are effective for improving token-level accuracy and specificity, they do not explicitly promote semantic coherence within the latent space.

To address this limitation, we introduce a latent-level alignment loss $\mathcal{L}_{\mathrm{LA}}$, which penalizes the discrepancy between the hidden representations of edited fact and its corresponding semantic anchors. Specifically, during the iterative optimization process, we compute the cosine distance between the hidden state $h_\delta^\ell$ (obtained from a forward pass with the current candidate vector $\delta$), and anchor $\varphi^\ell$ at selected mid-layers $\ell \in [\ell_{\text{start}}, \ell_{\text{end}}]$. The alignment loss $\mathcal{L}_{\mathrm{LA}}$ is defined as the average of these distances across layers:
\begin{equation}
    \mathcal{L}_{\mathrm{LA}}(\delta)
= \frac{1}{N_{\text{align}}}
\sum_{\ell' = \ell_{\text{start}}}^{\ell_{\text{end}}}
\Bigl[
1 - \cos(h_\delta^{\ell'}, \varphi^{\ell'})
\Bigr],
\end{equation}
\noindent where $N_{\text{align}}\!=\!\ell_{\text{end}}\!-\!\ell_{\text{start}}\!+\!1$ denotes the number of layers included in the computation. We incorporate $\mathcal{L}_{\mathrm{LA}}$ into the overall objective, yielding the following composite loss:
\begin{equation}
    \min_{\delta}\mathcal{L}(\delta)= \mathcal{L}_{\mathrm{NLL}}(\delta)+ \mathcal{L}_{\mathrm{KL}}(\delta)+ \lambda\mathcal{L}_{\mathrm{LA}}(\delta),
\end{equation}
where the hyperparameter $\lambda$ controls the weight of the latent-level alignment loss term. By minimizing $\mathcal{L}_{\mathrm{LA}}$, edited knowledge is more effectively integrated into the model’s latent semantic space and becomes organically connected to related knowledge, rather than exist as a standalone piece of information.

\begin{table*}[ht]
    \centering
    \small
    \renewcommand{\arraystretch}{1.3}
    \resizebox{\textwidth}{!}{
    \begin{tabular}{c l *{7}{c}}
        \toprule
        \multicolumn{1}{c}{\textbf{Model}}
          & \multicolumn{1}{c}{\textbf{Editor}}
          & \multicolumn{1}{c}{\textbf{Edit.}}
          & \multicolumn{1}{c}{\textbf{Effi.}}
          & \multicolumn{1}{c}{\textbf{Para.}}
          & \multicolumn{1}{c}{\textbf{Neigh.}}
          & \multicolumn{1}{c}{\textbf{Port.}}
          & \multicolumn{1}{c}{\textbf{Flu.}}
          & \multicolumn{1}{c}{\textbf{Cons.}} \\
        \midrule
        \multirow{4}{*}{\makecell[c]{\textbf{GPT-J}\\(6B)}}
        & ROME
          & 62.9 &100.0 & 99.5 & 79.1 & 32.4 & 619.7 & 46.4 \\
        & \textsc{Steam}$_\text{ROME}$
          & \textbf{67.1} {\scriptsize\textcolor{OliveGreen}{(+4.2)}} 
          &100.0 & 99.5 & 79.5 
          & \textbf{37.0} {\scriptsize\textcolor{OliveGreen}{(+4.6)}} 
          & 620.0 & \textbf{47.1} {\scriptsize\textcolor{OliveGreen}{(+0.7)}} \\
        & R-ROME
          & 62.7 &100.0 & 99.3 & 80.5 & 31.9 & 621.0 & 46.2 \\
        & \textsc{Steam}$_\text{R-ROME}$
          & \textbf{66.4} {\scriptsize\textcolor{OliveGreen}{(+3.7)}} 
          &100.0 & 99.4 & 80.5 
          & \textbf{36.0} {\scriptsize\textcolor{OliveGreen}{(+4.1)}} 
          & 620.6 & \textbf{47.0} {\scriptsize\textcolor{OliveGreen}{(+0.8)}} \\
        \midrule
        \multirow{4}{*}{\makecell[c]{\textbf{Qwen2}\\(7B)}}
        & ROME
          & 74.2 & 99.8 & 97.9 & 85.8 & 45.4 & 624.4 & 40.5 \\
        & \textsc{Steam}$_\text{ROME}$
          & \textbf{75.5} {\scriptsize\textcolor{OliveGreen}{(+1.3)}} 
          & 99.8 & 98.0 & 85.8 
          & \textbf{47.4} {\scriptsize\textcolor{OliveGreen}{(+2.0)}} 
          & 624.9 & \textbf{40.9} {\scriptsize\textcolor{OliveGreen}{(+0.4)}} \\
        & R-ROME
          & 73.4 & 99.0 & 95.8 & 86.2 & 44.8 & 624.5 & 39.7 \\
        & \textsc{Steam}$_\text{R-ROME}$
          & \textbf{75.3} {\scriptsize\textcolor{OliveGreen}{(+1.9)}} 
          & 99.7 & 96.8 & 86.1 
          & \textbf{47.3} {\scriptsize\textcolor{OliveGreen}{(+2.5)}} 
          & 625.1 & \textbf{40.4} {\scriptsize\textcolor{OliveGreen}{(+0.7)}} \\
        \midrule
        \multirow{4}{*}{\makecell[c]{\textbf{Llama3}\\(8B)}}
        & ROME
          & 72.6 &100.0 & 99.0 & 85.9 & 42.8 & 626.2 & 43.3 \\
        & \textsc{Steam}$_\text{ROME}$
          & \textbf{74.9} {\scriptsize\textcolor{OliveGreen}{(+2.3)}} 
          & 99.6 & 99.2 & 87.2 
          & \textbf{45.9} {\scriptsize\textcolor{OliveGreen}{(+3.1)}} 
          & 622.7 & 42.7 {\scriptsize\textcolor{Orange}{(-0.6)}} \\
        & R-ROME
          & 71.9 &100.0 & 98.4 & 86.9 & 41.7 & 625.4 & 42.4 \\
        & \textsc{Steam}$_\text{R-ROME}$
          & \textbf{76.0} {\scriptsize\textcolor{OliveGreen}{(+4.1)}} 
          & 99.9 & 99.1 & 87.5 
          & \textbf{47.4} {\scriptsize\textcolor{OliveGreen}{(+5.7)}} 
          & 622.8 & 42.2 {\scriptsize\textcolor{Orange}{(-0.2)}} \\
        \bottomrule
    \end{tabular}
    }
    \caption{Single-editing results on \textsc{CounterFactPlus} for GPT-J (6B), Qwen2 (7B), and Llama3 (8B). Results are averaged over three runs. Values in parentheses indicate the improvement or decline relative to the baseline for Edit score, Portability, and Consistency.}
    \label{tab:single_editing}
\end{table*}

\section{Experiments}\label{sec:5}
\subsection{Experimental Settings}
\textbf{Baseline Models.} We evaluate our approach using GPT-J (6B)~\cite{gpt-j}, Qwen2 (7B)~\cite{yang2024qwen2technicalreport}, Llama 3 (8B)~\cite{dubey2024llama} to ensure the robustness of our findings.
\newline\noindent\textbf{Dataset.} We utilize the \textsc{CounterFactPlus}~\cite{CounterFactPlus} dataset, which contains more challenging questions that require models to reason with edited knowledge. To construct robust semantic anchors, we exclude samples where the filtered reference set $T'$ contains fewer than 32 triples. Consequently, from the initial dataset of 1,031 samples, we utilized 816 for GPT-J, 927 for Qwen2, and 937 for Llama 3.
\newline\noindent\textbf{Knowledge editing approaches.} We evaluate our method under two scenarios: (1) \textit{\textbf{Single editing}}, where each factual update is applied independently (i.e., the model is restored after each edit), and (2) \textit{\textbf{Batch editing}}, where multiple edits are applied simultaneously. For single editing, we apply our framework to existing methods, including ROME~\cite{ROME} and R-ROME~\cite{R-ROME}. For batch editing, we integrate our method with PMET~\cite{PMET}, a state-of-the-art batch editing method. Implementation details are provided in Appendix~\ref{appendix:implementation_details}.
\newline\noindent\textbf{Metrics.}
We evaluate all methods using several metrics: \textit{\textbf{Efficacy Score}} (Effi.) measures whether the model accurately recalls the updated fact; \textit{\textbf{Paraphrase Score}} (Para.) assesses the generalizability of the edit; \textit{\textbf{Neighborhood Score}} (Neigh.) evaluates whether unrelated knowledge remains unaffected; and \textit{\textbf{Portability Score}} (Port.) measures the model’s ability to apply the edited knowledge in multi-hop reasoning. \textit{\textbf{Edit Score}} (Edit.) is the harmonic mean of these four metrics. Additionally, we report \textit{\textbf{Fluency}} (Flu.), which reflects lexical diversity of generated text, 
and \textit{\textbf{Consistency}} (Cons.), which captures semantic coherence.

\subsection{Results}\label{sec:5_2}
\subsubsection{Single Editing}\label{sec:5_2_1}
In the single editing scenario, we evaluate our \textsc{Steam} framework by integrating it into both ROME and R-ROME, denoted as \textsc{Steam}$_{\text{ROME}}$ and \textsc{Steam}$_{\text{R-ROME}}$, respectively. For all models, we set $\ell_{\text{start}} = 13$, $\ell_{\text{end}} = 17$, and $\lambda = 5$, and provide a detailed hyperparameter analysis in Appendix~\ref{appendix:hyperparam}.

As summarized in Table~\ref{tab:single_editing}, \textsc{Steam} demonstrates consistent improvements in editing quality across all evaluated models. Portability increases robustly in all cases, with the highest gain of +5.7 observed in Llama3 under \textsc{Stake}$_{\text{R-ROME}}$, indicating that edited knowledge is more effectively integrated and applied in multi-hop reasoning. Importantly, these gains are observed consistently across both baseline editors, ROME and R-ROME, underscoring the generality of our framework.

Other metrics such as Efficacy, Paraphrase, and Neighborhood remain stable, with slight gains in some cases, showing that \textsc{Steam} preserves the local accuracy and generalization ability of the baseline editors. These steady or improved results collectively contribute to an overall enhancement of the aggregated Edit score. Notably, GPT-J and Qwen2 show modest but consistent improvements in Consistency (up to +0.8 and +0.7, respectively), highlighting \textsc{Steam}’s ability to enhance semantic coherence in edited knowledge.

% Interestingly, on Llama3, \textsc{Stake}$_{\text{R-ROME}}$ yields the largest Portability gain (+5.7) among all settings, but this comes with a slight decline in Consistency. This contrast suggests that while \textsc{Steam} strongly enhances the transferability of edited knowledge, its effect on maintaining semantic coherence may vary across model architectures.

\subsubsection{Batch Editing}\label{sec:5_2_2}
\begin{table*}[ht]
    \centering
    \small
    \renewcommand{\arraystretch}{1.4}
    \resizebox{\textwidth}{!}{
    % 10 columns: Model | Edits | Editor | Edit. | Effi. | Para. | Neigh. | Port. | Flu. | Cons.
    \begin{tabular}{c c l *{7}{c}}
        \toprule
        \textbf{Model} & \textbf{Edits} & \textbf{Editor}
          & \textbf{Edit.} & \textbf{Effi.} & \textbf{Para.}
          & \textbf{Neigh.} & \textbf{Port.} & \textbf{Flu.} & \textbf{Cons.} \\
        \midrule
        \multirow{8}{*}{\makecell{\textbf{GPT-J}\\(6B)}}
          & \multirow{2}{*}{1}
            & PMET
            & 63.5 & 100.0 & 97.0 & 82.2 & 32.8 & 620.7 & 45.3 \\
          &  & \textsc{Steam}$_{\text{PMET}}$
            & \textbf{65.5} {\scriptsize(\textcolor{OliveGreen}{+2.0})}
            & 100.0 & 96.6 & 82.4
            & \textbf{35.0} {\scriptsize(\textcolor{OliveGreen}{+2.2})}
            & 619.0 & \textbf{46.4} {\scriptsize(\textcolor{OliveGreen}{+1.1})} \\
        \arrayrulecolor{gray!50}\cdashline{2-10}\arrayrulecolor{black}
          & \multirow{2}{*}{10}
            & PMET
            & 65.0 & 100.0 & 97.2 & 82.2 & 34.4 & 620.6 & 45.2 \\
          &  & \textsc{Steam}$_{\text{PMET}}$
            & \textbf{65.9} {\scriptsize(\textcolor{OliveGreen}{+0.8})}
            & 100.0 & 96.6 & 82.4
            & \textbf{35.4} {\scriptsize(\textcolor{OliveGreen}{+1.0})}
            & 618.9 & \textbf{46.3} {\scriptsize(\textcolor{OliveGreen}{+1.1})} \\
        \arrayrulecolor{gray!50}\cdashline{2-10}\arrayrulecolor{black}
          & \multirow{2}{*}{100}
            & PMET
            & 64.1 &  99.9 & 96.9 & 81.6 & 33.5 & 621.3 & 45.3 \\
          &  & \textsc{Steam}$_{\text{PMET}}$
            & \textbf{66.0} {\scriptsize(\textcolor{OliveGreen}{+2.0})}
            &  99.9 & 96.6 & 81.9
            & \textbf{35.7} {\scriptsize(\textcolor{OliveGreen}{+2.2})}
            & 619.2 & \textbf{46.3} {\scriptsize(\textcolor{OliveGreen}{+1.0})} \\
        \arrayrulecolor{gray!50}\cdashline{2-10}\arrayrulecolor{black}
          & \multirow{2}{*}{All}
            & PMET
            & 63.7 & 100.0 & 96.4 & 77.2 & 33.9 & 621.5 & 45.0 \\
          &  & \textsc{Steam}$_{\text{PMET}}$
            & \textbf{65.0} {\scriptsize(\textcolor{OliveGreen}{+1.3})}
            & 100.0 & 96.0 & 77.2
            & \textbf{35.5} {\scriptsize(\textcolor{OliveGreen}{+1.6})}
            & 619.1 & \textbf{46.1} {\scriptsize(\textcolor{OliveGreen}{+1.1})} \\
        \bottomrule
    \end{tabular}
    }
    \caption{Batch-editing results on \textsc{CounterFactPlus} for GPT-J (6B). Results are averaged over three runs. Values in parentheses indicate improvements of \textsc{Steam}${_\text{PMET}}$ over the PMET baseline for Edit., Portability (Port.), and Consistency (Cons.). Each row reports performance at different batch sizes (1, 10, 100, and all 816 edits).}
    \label{tab:pmet_vs_steampmet}
\end{table*}

In the batch-editing scenario, we incorporate our \textsc{Steam} framework into PMET, denoted as \textsc{Steam}$_{\text{PMET}}$, and evaluate it on GPT-J. Experiments are conducted with batch sizes of 1, 10, 100, and all 816 edits in the dataset, as reported in Table~\ref{tab:pmet_vs_steampmet}.

Across all batch sizes, \textsc{Steam}$_{\text{PMET}}$ consistently improves Portability (up to +2.2), indicating that edited knowledge is more effectively transferred and applied even under large-scale editing. Edit scores also increase across settings (up to +2.0), suggesting that the overall quality of edits benefits from semantic alignment. Consistency further exhibits modest but reliable gains (up to +1.1), highlighting that \textsc{Steam} helps preserve semantic coherence in batch-editing conditions. Meanwhile, other metrics such as Efficacy, Paraphrase, and Neighborhood remain stable, showing that local factual accuracy and generalization are not sacrificed. Collectively, these results demonstrate that \textsc{Steam} provides robust improvements in knowledge integration across varying batch sizes.

\subsection{Discussion}\label{sec:5_3}
\textbf{Verifying Latent-Space Alignment of Edited Knowledge.} To assess whether the performance improvements achieved by \textsc{Steam} stem from latent-level integration, we revisit the latent space visualizations. As shown in Figure~\ref{fig:overview_rome_vs_stake}(b), which illustrates the residual streams produced by \textsc{Steam}$_\text{ROME}$, the hidden states of the edited knowledge (red diamonds) exhibit trajectories that are more closely aligned with those of the reference knowledge (green circles), compared to the baseline ROME shown in Figure~\ref{fig:overview_rome_vs_stake}(a). This alignment suggests that \textsc{Steam} effectively guides the representation of edited facts toward semantically coherent regions within the model’s latent space. These findings indicate that the observed performance gains are the result of genuine semantic-level integration, rather than surface-level adjustments. To further support this interpretation, we conduct a layer-wise cosine similarity analysis (Appendix~\ref{appendix:cosine_sim_analysis}), which quantitatively confirms stronger alignment between the edited representations and their corresponding semantic anchors.

\section{Conclusion}
In this paper, we addressed a limitation of current knowledge editing methods: their failure to coherently integrate updated knowledge into the model's existing structure despite successful output changes. Our analysis revealed that existing methods often induce isolated latent representations, hindering consistent and reliable inference with edited knowledge. To address this issue, we introduced \textsc{Steam}, a framework designed to promote deeper semantic integration. Our experimental results demonstrate that enhancing semantic alignment improves the model's reasoning capabilities with edited knowledge and increases semantic coherence. These findings emphasize the importance of semantic-level integration for developing more robust and coherent knowledge editing techniques.

\clearpage
\section*{Limitation}
\textbf{Dependence on External Knowledge.} \textsc{Steam} constructs semantic anchors based on reference facts retrieved from external structured knowledge sources such as Wikidata. This design enables the model to leverage well-grounded factual associations when guiding semantic alignment. Therefore, for newly emerging or less well-known entities, retrieving relevant reference knowledge may be difficult, which can limit the applicability of the proposed framework.

\noindent\textbf{Anchor Construction and Knowledge Selection.} \textsc{Steam}  assumes that the latent representation of an updated fact can be approximated by aggregating reference knowledge about the same object. While this offers a practical signal for alignment, it may not fully reflect how language models internally structure and reason over knowledge. This process in closely related to broader questions about the structure of factual reasoning in language models\,--\,a topic that remains underexplored. We view our method as a first step in this direction and leave more systematic strategies for anchor construction and deeper investigation of knowledge inference mechanisms for future work.

\section*{Acknowledgements}
This work was supported by the National Research Foundation of Korea (NRF) grant funded by the Korea government (MSIT) (RS-2025-00553041, Enhancement of Rational and Emotional Intelligence of Large Language Models for Implementing Dependable Conversational Agents) and the Institute for Information \& communications Technology Promotion (IITP) grant funded by the Korea government (MSIT) (RS-2024-00398115, Research on the reliability and coherence of outcomes produced by Generative AI).

\bibliography{custom}

\clearpage  % 페이지 넘기기 추가함
\appendix
\section*{Appendix}
\section{Quantitative Latent-Space Analysis via Cosine Similarity}\label{sec:appendix_visualization}
To complement the qualitative findings presented in Section~\ref{sec:3_2}, we provide additional visualization results for both GPT-J and GPT-2 XL~\cite{gpt-xl}. These visualizations illustrate how edited knowledge propagates through the residual stream and how it aligns (or diverges) from reference knowledge across transformer layers.

\begin{figure*}[t]
\includegraphics[width=\textwidth]{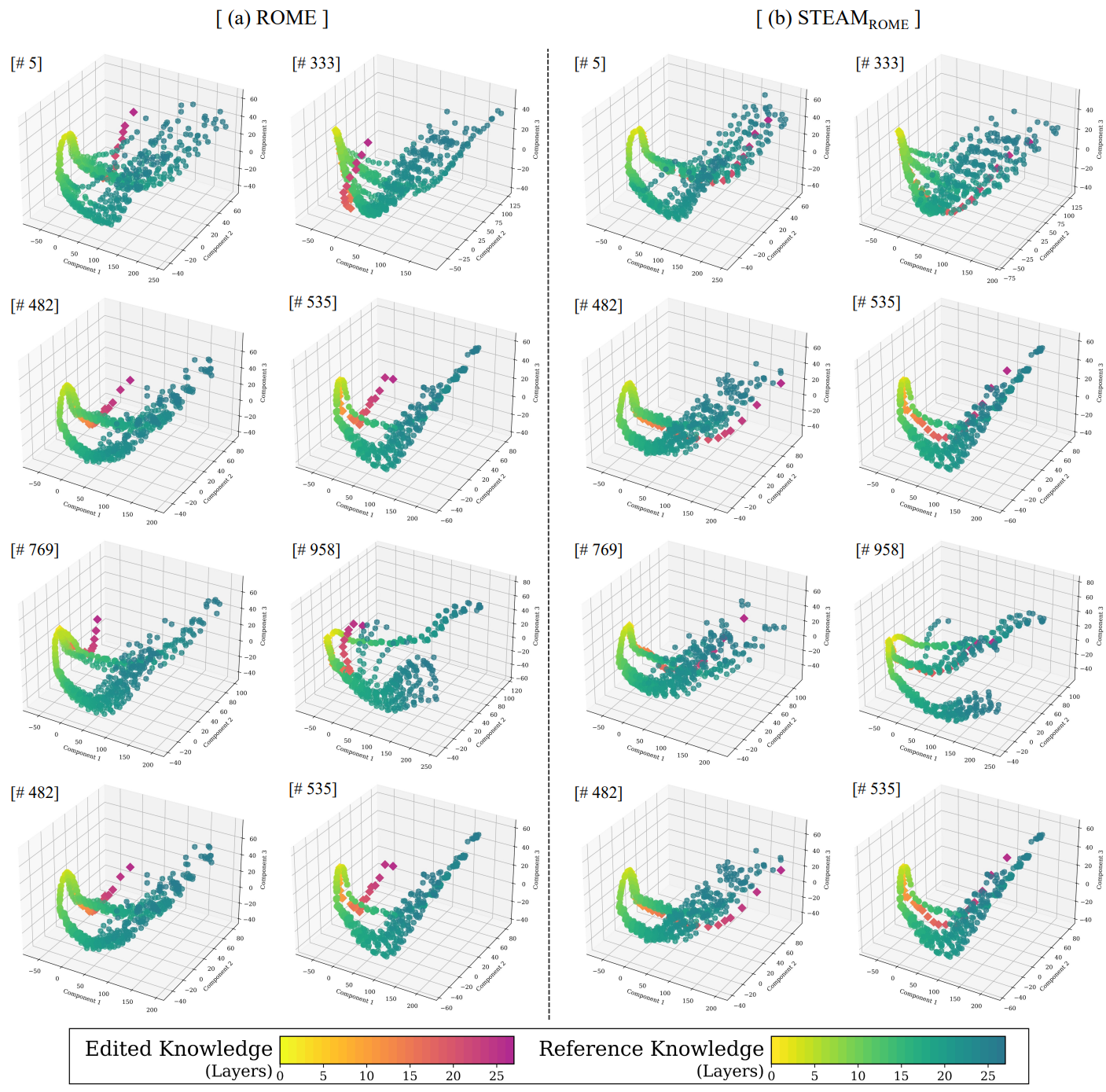}
  \caption{3D visualization of residual stream representations across layers for GPT-J. Each subplot shows PCA-projected hidden states of an edited fact (red diamonds) and its corresponding reference facts (green circles), with sample indices indicated in brackets. (a) shows results under ROME, and (b) under \textsc{Steam}$_\text{ROME}$.}
  \label{fig:GPT-J_additional}
\end{figure*}

\begin{figure*}[t]
\includegraphics[width=\textwidth]{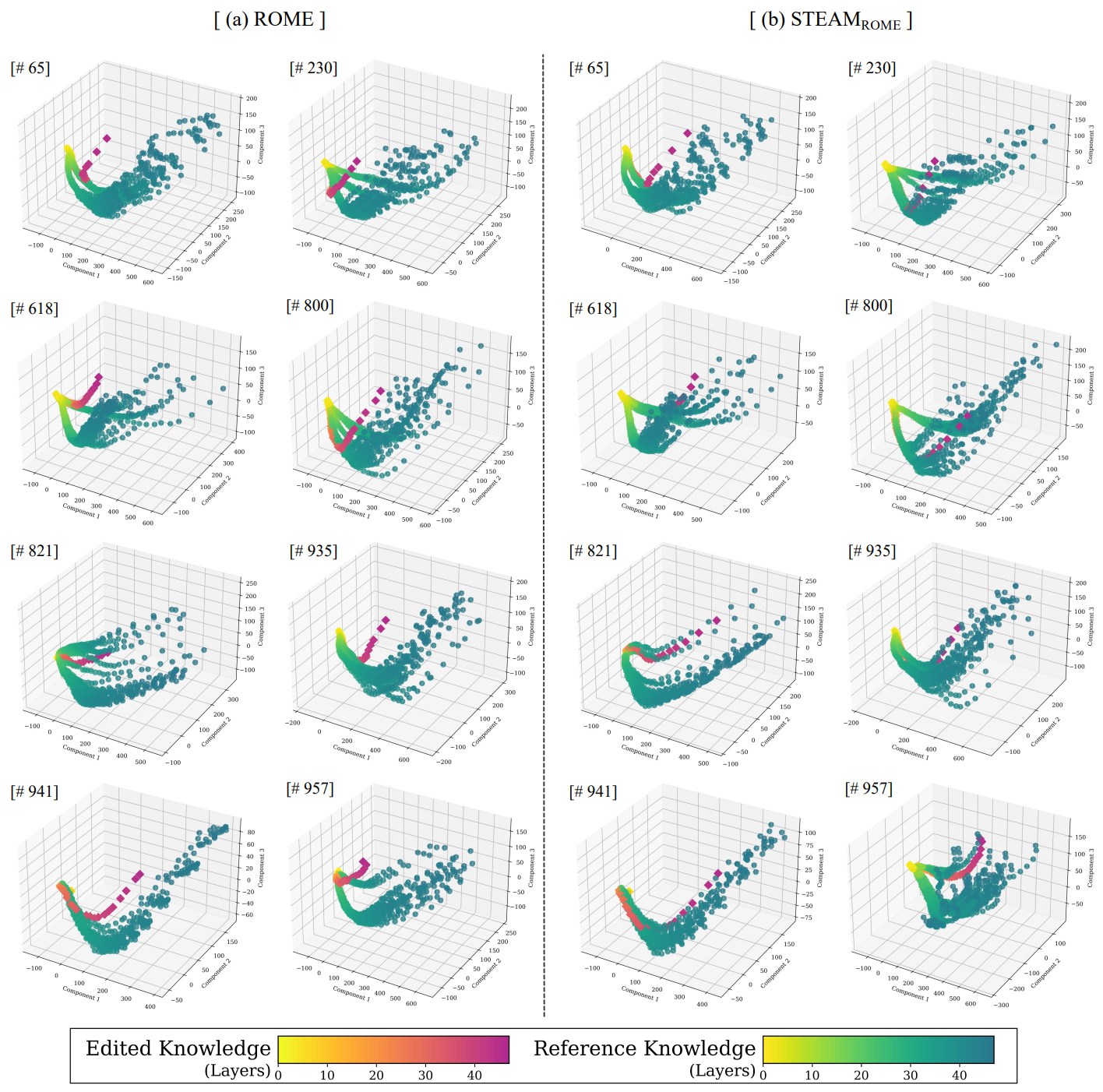}
  \caption{3D visualization of residual stream representations across layers for GPT2-XL. Each subplot shows PCA-projected hidden states of an edited fact (red diamonds) and its corresponding reference facts (green circles), with sample indices indicated in brackets. (a) shows results under ROME, and (b) under \textsc{Steam}$_\text{ROME}$.}
  \label{fig:GPT2-XL_additional}
\end{figure*}

Figure~\ref{fig:GPT-J_additional} presents PCA-projected hidden states for multiple samples from the \textsc{CounterFact} dataset. Red diamonds represent the residual stream of the edited knowledge, while green circles correspond to the reference knowledge recalling the same target object $o^*$. Under baseline editing methods (ROME), the edited knowledge forms a semantically isolated stream in the latent space, following a path that deviates from reference knowledge. In contrast, under \textsc{Steam}-enhanced editing, the edited trajectories are more closely aligned with reference knowledge, indicating improved semantic integration. Figure~\ref{fig:GPT2-XL_additional} provides corresponding results for GPT-2 XL. Despite architectural differences and increased depth, we observe a similar trend, reinforcing the model-agnostic effectiveness of \textsc{Steam}. These qualitative observations, consistent across models and examples, provide strong visual evidence that \textsc{Steam} facilitates meaningful representational alignment of edited knowledge within the latent space.

\section{Measuring Semantic Alignment}\label{appendix:cosine_sim_analysis}
\begin{figure*}[t]
\includegraphics[width=\linewidth]{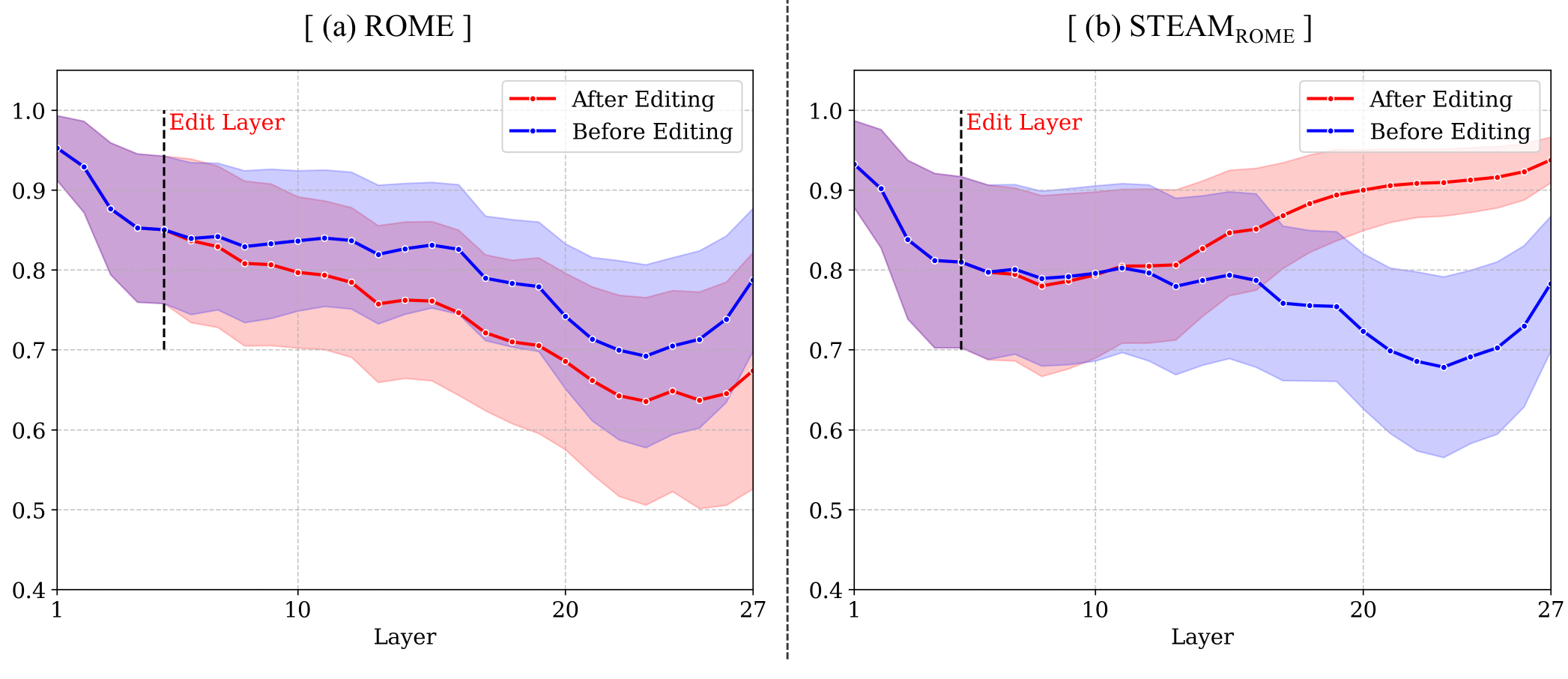}
  \caption{Layer-wise cosine similarity between model representations and semantic anchors in GPT-J. Each plot shows the average cosine similarity between anchor vectors $\varphi^\ell$ and hidden states from the edited model $h_{\varepsilon}^\ell$ (red) and the unedited model $h_\theta^\ell$ (blue), with shaded areas indicating standard deviation. The vertical dashed line marks the edit layer. (a) Result with ROME. (b) Result with \textsc{Steam}$_\text{ROME}$.}
  \label{fig:cosine}
\end{figure*}
While existing editing methods can successfully update model outputs, it remains unclear whether such updates are semantically integrated into the model’s internal knowledge structure. To investigate this, we conduct a layer-wise cosine similarity analysis between the edited knowledge representation and a reference semantic anchor vector.

For each edit $(s, r, o \rightarrow o^*)$, we first construct a baseline semantic representation $\{\varphi^\ell\}_{\ell=1}^L$ for the updated object $o^*$ by aggregating the hidden states of reference facts, as described in Section~\ref{sec:4_1}. We then compare how the edited model $\mathcal{F}'$ represents $o^*$ in response to the edited prompt $p_\varepsilon$ (e.g., \texttt{Eiffel Tower stands in \underline{\hspace{0.4cm}}} ) by extracting the hidden states $\{h_\varepsilon^\ell\}_{\ell=1}^L$ and computing:
$$
\mathcal{S}(\varphi, h_\varepsilon) = \left\{ \cos\left( \varphi^\ell, h_\varepsilon^\ell \right) \right\}_{\ell=1}^L,
$$
where $\text{cos}(\cdot,\cdot)$ denotes the cosine similarity. We also apply the same procedure to the unedited model $\mathcal{F}$, obtaining $\mathcal{S}(\varphi, h_{\theta})$. Since $(s,r)\to o$ (e.g., \texttt{Eiffel Tower stands in} → \texttt{Paris}) and $(s_i, r_i) \to o^*$ (e.g., \texttt{River Thames flows through} $\rightarrow$ \texttt{London}) are semantically unrelated facts, this comparison provides a reference for assessing the inherent similarity between different knowledge representations in the model.

Figure~\ref{fig:cosine} (a) presents the layer-wise cosine similarity scores for GPT-J under the ROME baseline. Where the red curve represents $S\bigl(\varphi, h_{\varepsilon}\bigr)$ for the edited model and the blue curve represents $S\bigl(\varphi, h_{\theta}\bigr)$ for the base model. Up to the edit layer, both curves are the same; after that, they diverge, showing a representational shift induced by the factual update. However, this divergence does not necessarily indicate that the updated knowledge has been meaningfully integrated. Instead, as the layers progress, the red curve follows a pattern similar to the blue curve while showing a gradual decline. Given the distinct semantics of $o$ and $o^*$ (e.g., \texttt{Paris} vs. \texttt{London}), these findings suggest that although the edit does change the updated fact’s representation, the new knowledge remains separate from the original latent structure.

Figure~\ref{fig:cosine} (b) shows the corresponding analysis for the \textsc{Steam}-enhanced model. In contrast to the baseline, the red curve (after editing) exhibits a clear upward trend after the edit layer, diverging from the blue curve (before editing) and aligning more closely with the semantic anchor $\varphi^\ell$. This indicates that the updated knowledge $(s, r) \to o^*$ is not only distinguishable from the original fact $(s, r) \to o$, but is also actively aligned with the reference knowledge encoding of $o^*$. These results quantitatively validate the alignment mechanism introduced in our framework and support the interpretability of \textsc{Steam}’s improvements in semantic coherence.
\section{Reference Triple Selection and Filtering Criteria}\label{sec:appendix_reference_triple}
\begin{algorithm}[htb]
\caption{Stratified Sampling by Relation Type}
\label{alg:stratified_sample}
\begin{algorithmic}[1]
\Require Dataset $T$, Sample size $N$
\Ensure Stratified sample $T'$ 

\Statex \footnotesize \texttt{/* Check if dataset is empty */} \normalsize
\If{$T$ is empty}
    \State \Return $\emptyset$
\EndIf

\Statex \footnotesize \texttt{/* Group dataset by relation type */} \normalsize
\State $G \gets$ group $T$ by relation type
\State $R \gets |G|$  

\Statex \footnotesize \texttt{/* Compute base and extra samples */} \normalsize
\State $n \gets \lfloor \frac{N}{R} \rfloor$  
\State $r \gets N - (n \times R)$  

\Statex \footnotesize \texttt{/* Initialize sampled dataset */} \normalsize
\State $T' \gets \emptyset$

\Statex \footnotesize \texttt{/* Stratified sampling per group */} \normalsize
\ForAll{$g \in G$}
    \State $s \gets n$
    \If{$r > 0$}
        \State $p \gets \frac{|g|}{|T|}$  
        \State $a \gets \lfloor r \times p \rfloor$  
        \State $s \gets s + a$
    \EndIf
    \State $T' \gets T' \cup$  
    \Statex \hfill \textsc{RandomSample}$(g, \min(s, |g|))$
\EndFor

\Statex \footnotesize \texttt{/* Handle remaining samples if needed */} \normalsize
\If{$|T'| < N$}
    \State $U \gets T \setminus T'$
    \State $m \gets N - |T'|$
    \State $T' \gets T' \cup$  
    \Statex \hfill \textsc{RandomSample}$(U, \min(m, |U|))$
\EndIf

\Statex \footnotesize \texttt{/* Return final sampled dataset */} \normalsize
\State \Return $T'$

\end{algorithmic}
\end{algorithm}
For each editing sample $\varepsilon$, we begin by collecting a set of reference triples $T=\{(s_i, r_i, o^*)\}_{i=1}^N$. We then transform each triple into a textual prompt $p_i$. Following previous work~\cite{ClozePrompt}, we define a template $t_{r_i}$ for each relation $r_i$. By inserting the subject $s_i$ into this template, we generate a textual prompt $p_i=t_{r_i}(s_i)$. For instance, if $s_i$ is “United States” and $r_i$ is “capital of”, then $p_i$ is “The capital of United States is”, which is designed to elicit $o^*$. Next, we input $p_i$ into the unedited model $\mathcal{F}$, performing greedy decoding for $k$ subword tokens, where $k$ corresponds the subword length of $o^*$. Let $\mathrm{dec}(\mathcal{F}(p_i), k)$ denote the top-ranked sequence of length $k$. If this sequence exactly matches $o^*$, we infer that the model recognizes this fact. Formally, the filtered set $T'\subseteq T$ is
\begin{equation}
T'\!=\!\Bigl\{(s_i, r_i, o^*)\!\in\! T\,\Big|\,
\mathrm{dec}\bigl(\mathcal{F}(p_i),\,k\bigr)\!=\!o^*\!\Bigr\}.
\label{eq:triple_inference}
\end{equation}

To construct a balanced anchor set, we apply a stratified sampling procedure to the filtered set $T'$. Specifically, we group all triples in $T'$ by their relation type $r_i$, and assign an equal base number of samples to each group. Any remaining quota is distributed proportionally based on group sizes. If a group contains fewer examples than its assigned quota, we include all available triples from that group. Finally, if the total number of selected samples is still below the target size, we randomly sample additional instances from the remaining pool to complete the set. The full procedure is detailed in Algorithm \ref{alg:stratified_sample}. This sampling strategy ensures that the resulting subset $T'' \subseteq T'$ reflects a diverse and representative distribution of relation types, thereby enabling the construction of robust and unbiased semantic anchors. In our implementation, we set the maximum size of $T''$ to 64 reference triples per edit.

\section{\textsc{CounterFactPlus}}\label{sec:appendix_dataset}

\begin{table*}[htb!]
  \centering
  \resizebox{\linewidth}{!}
  {%
  \begin{tabularx}{\linewidth}{lX} 
    \toprule
    \textbf{Property} & \textbf{Value} \\
    \midrule
    Edit request ($\varepsilon$) & (Skype, product of, Microsoft $\to$ Apple) \\
      Efficacy prompt ($P^E$) & Skype was a product of \\
      Paraphrase prompt ($P^P$) & He moved to Edmonton, Alberta, when he was seven-years old. Skype was created by \\
      Neighborhood prompt ($P^N$) & Windows Media Center, a product created by \\
      Recalled knowledge ($o^*, r', o'$) & (Apple, founded by, Steve Jobs and Steve Wozniak) \\
      Multi-hop question ($q'$) & Who are the founders of the company that created Skype? \\
      Multi-hop Answer ($o'$)& Steve Jobs and Steve Wozniak \\
    \bottomrule
  \end{tabularx}
  }
  \caption{An Example of \textsc{CounterFactPlus} Dataset}
  \label{cf_sample}
\end{table*}

The \textsc{CounterFactPlus} dataset is an enhanced version of the original \textsc{CounterFact} benchmark, comprising 1,031 selected entries from the original data. 
Table \ref{cf_sample} presents an example from the \textsc{CounterFactPlus} dataset. Each sample includes a factual edit $\varepsilon$, such as (Skype, product of, Microsoft $\rightarrow$ Apple), diverse set of prompts that assess the impact of the edit from multiple perspectives.
\begin{itemize}[leftmargin=8pt, topsep=6pt]
\setlength\itemsep{0em}
    \item \textbf{Efficacy Prompt} ($P^E$) is used to assess whether the edited knowledge is accurately reflected in the model’s output.
    \item \textbf{Paraphrase Prompt} ($P^P$) tests whether the edited information is consistently maintained across different surface forms.
    \item \textbf{Neighborhood Prompt} ($P^N$) provides unrelated factual contexts and is used to evaluate whether the model preserves non-target knowledge after editing.
    \item \textbf{Multi-hop Question} ($q'$) assesses whether the model can reason over multiple connected facts involving the updated knowledge. Specifically, each question requires the model to combine the edited fact $(s, r, o^*)$ with an additional fact $(o^*, r', o')$, and infer the answer $o'$ based on this reasoning path. This evaluates whether the updated object $o^*$ has been semantically integrated into the model’s broader knowledge structure in a way that supports multi-step inference.
\end{itemize}
\section{Implementation Details} \label{appendix:implementation_details}
Our single-editing experiments were conducted on GPT-J (6B), Qwen2 (7B), and Llama3 (8B), while batch-editing experiments were performed only on GPT-J (6B). All experiments were run on an NVIDIA A100 (80GB) GPU using the Adam optimizer.  

For ROME and R-ROME, the primary editing layer is set to 5 across all models, based on prior findings about where factual associations are stored. The value vector $v^*$ is optimized for 20 gradient steps with a learning rate of 0.5 and a weight decay of 0.5. The loss is computed at layer 27, and the KL divergence regularization factor is set to 0.0625.  

For PMET (batch editing), we follow the original configuration and apply edits across a broader range of layers [3, 4, 5, 6, 7, 8] on GPT-J. We set the number of gradient steps to 30, the learning rate to 0.2, and the clamp normalization factor to 0.75. The KL regularization factor is set to 1.0, and the loss is computed at layer 27.
\section{Metrics}\label{sec:appendix_metrics}

We evaluate the performance of knowledge editing methods using six metrics that assess factual accuracy, generalization, locality, and semantic coherence. These metrics are computed over samples from the \textsc{CounterFactPlus} dataset and are defined as follows:
\begin{itemize}[leftmargin=8pt, topsep=6pt]
\setlength\itemsep{0em}
    \item \textbf{Efficacy Score} evaluates whether the edited model $\mathcal{F}'$ correctly recalls the updated object $o^*$ given the edit prompt $p \in P^E$. It is computed as:
    \begin{equation*}
        \mathbb{E}_{p \in P^E}\left[\mathbb{\mathbb { I }}\left[\mathbb{P}_\mathcal{F'}\left(o^{*}\mid p\right)>\mathbb{P}_\mathcal{F'}\left(o\mid p\right)\right]\right],
    \end{equation*}
    where $\mathbb{E}$ denotes the average, $\mathbb{I}$ denotes the indicator function, and $\mathcal{F'}$ represents the edited model. $\mathbb{P}(o\mid p)$ indicates the probability that the model generates output $o$ given the input prompt $p$.
    \item \textbf{Paraphrase Score} measures whether the model maintains the updated knowledge under paraphrased prompts $p \in P^P$. The formulation mirrors that of the Efficacy Score:
    \begin{equation*}
        \mathbb{E}_{p \in P^P}\left[\mathbb{\mathbb { I }}\left[\mathbb{P}_\mathcal{F'}\left(o^{*}\mid p\right)>\mathbb{P}_\mathcal{F'}\left(o\mid p\right)\right]\right].
    \end{equation*}
    \item \textbf{Neighborhood Score} assesses whether the model preserves unrelated knowledge after editing. For neighborhood prompts $p \in P^N$, which should not reflect the updated fact, the metric is computed as:
    \begin{equation*}
        \mathbb{E}_{p \in P^N}\left[\mathbb{\mathbb { I }}\left[\mathbb{P}_\mathcal{F'}\left(o^{*}\mid p\right)>\mathbb{P}_\mathcal{F'}\left(o\mid p\right)\right]\right].
    \end{equation*}
    \item \textbf{Portability Score} evaluates whether the model can correctly answer multi-hop questions that require reasoning over the edited knowledge. Each question $q'$ is constructed to test whether the model can infer the final object $o'$ by combining the edited fact $(s, r, o^*)$ with an additional fact $(o^*, r’, o')$. The model’s response $\mathcal{F'}(q')$ is evaluated against the gold answer $o'$ using fuzzy string matching. A prediction is considered correct if the partial ratio ($PR$) between the model’s output and $o'$ exceeds 0.7:
    \begin{equation*}
        \mathbb{E}_{q',o' \in Q}\left[\mathbb{\mathbb {I}}\left[PR(\mathcal{F'}(q'),o')> 0.7\right]\right].
    \end{equation*}
    where $PR$ denotes the partial ratio from the Fuzzywuzzy library\footnote{https://github.com/seatgeek/fuzzywuzzy}, based on Levenshtein distance.
    \item \textbf{Fluency} measures evaluates the repetitiveness of generated text by computing the weighted average entropy over bi-grams and tri-grams, following the method proposed by~\citealp{Fluency_ref}. Formally, it is calculated as:
    \begin{equation*}
        \sum_k{f(k)log_2f(k)}
    \end{equation*}
    where $f(k)$ denotes the frequency distribution of n-grams. Lower values indicate repetitive or less diverse outputs, while higher values reflect more natural and varied language generation.
    \item \textbf{Consistency} measures the semantic consistency of the generated outputs. To compute this score, we generate text from a generation prompt $p \in P^G$, and calculate the cosine similarity between the TF-IDF vectors of the generated text and reference texts($t_{ref}$) about subjects sharing the target property $o^*$. Formally, it is calculated as:
    \begin{equation*}
        \mathbb{E}_{p \in P^G}[cos(\mathcal{F'}(p),t_{ref})]
    \end{equation*}
\end{itemize}
\section{Analyzing the Impact of Alignment Strength and Layer Choice}\label{appendix:hyperparam}
To better understand how the latent-level alignment in \textsc{Steam} affects the integration of edited knowledge, we analyze two key factors: the alignment strength ($\lambda$) and the depth of the alignment layer range. All experiments in this analysis are conducted on GPT-J (6B), using \textsc{Steam}$_\text{ROME}$.

\noindent\textbf{Effect of Alignment Strength.} We first fix the alignment layers to $\ell_{\text{start}}=13$ and $\ell_{\text{end}}=18$, and vary the alignment strength $\lambda \in {1, 3, 5, 10}$. Results show that moderate increases in $\lambda$ improve semantic integration. For example, Portability improves from 34.7 ($\lambda=1$) to 37.0 ($\lambda=5$), and Consistency increases from 47.0 to 47.2. However, when $\lambda$ is further increased to 10, these gains do not continue to rise and begin to plateau or slightly decline. Additionally, Fluency decreases steadily with stronger alignment (620.7 → 619.6 → 618.5), suggesting that overly strong alignment may introduce artifacts or reduce generation quality. These results imply that while stronger alignment encourages integration of edited knowledge, there exists an optimal range beyond which further gains are marginal or even detrimental.

\noindent\textbf{Effect of Alignment Layer Depth.} Next, we fix $\lambda=5$ and vary the alignment layers across three ranges: [8–13], [13–18], and [18–23]. We observe that deeper alignment layers generally yield higher Portability (35.0 → 35.7 → 37.9), while Consistency remains stable (46.8 → 47.2). However, Fluency shows a steady decline (621.1 → 619.6 → 618.5) as the alignment moves to later layers. These findings suggest that deeper layers, while capable of capturing more abstract semantics, may lack representational stability, making them less ideal as anchor points. Mid-layer ranges thus offer a favorable balance\,--\,rich enough in semantic information while still providing stable targets for alignment.
\clearpage

\clearpage
\clearpage

\end{document}